\documentclass[12pt]{iopart}

\usepackage{cite}
\usepackage{amsmath,amssymb,amsfonts}

\usepackage{graphicx}
\usepackage{textcomp}
\usepackage{xcolor}
\usepackage{adjustbox}
\usepackage{multirow}
\usepackage[numbers,sort&compress]{natbib}
\usepackage{bm}
\usepackage{array}

\usepackage{url}
\usepackage{appendix}

\usepackage{algpseudocode}
\usepackage{tikz}
\newcommand{\re}[1]{\textcolor{black}{#1}}

\begin{document}

\title[Embedding Decomposition]{Embedding Decomposition for Artifacts Removal in EEG Signals}

\author{Junjie Yu$^1$, Chenyi Li$^{1,2}$,
Kexin Lou$^1$, Chen Wei$^1$ and Quanying Liu$^{1,*}$}

\address{$^1$ Shenzhen Key Laboratory of Smart Healthcare Engineering, Department of Biomedical Engineering, Southern University of Science and Technology, Shenzhen 518055, People's Republic of China}
\address{$^2$ School of Science and Engineering, The Chinese University of Hong Kong (Shenzhen), Shenzhen 518172, People's Republic of China}
\address{$^*$ Author to whom any correspondence should be addressed}
\ead{liuqy@sustech.edu.cn}


\begin{abstract}
\textit{Objective.}
Electroencephalogram (EEG) recordings are often contaminated with artifacts. 
Various methods have been developed to eliminate or weaken the influence of artifacts. 
However, most of them rely on prior experience for analysis.
\textit{Approach.}
Here, we propose an deep learning framework to separate neural signal and artifacts in the embedding space and reconstruct the denoised signal, which is called DeepSeparator.
DeepSeparator employs an encoder to extract and amplify the features in the raw EEG, a module called decomposer to extract the trend, detect and suppress artifact and a decoder to reconstruct the denoised signal.
Besides, DeepSeparator can extract the artifact, which largely increases the model interpretability. 
\textit{Main results.}
The proposed method is tested with a semi-synthetic EEG dataset and a real task-related EEG dataset, suggesting that DeepSeparator outperforms the conventional models in both EOG and EMG artifact removal.
\textit{Significance.}
DeepSeparator can be extended to multi-channel EEG and data with any arbitrary length. It may motivate future developments and application of deep learning-based EEG denoising. The code for DeepSeparator is available at \url{https://github.com/ncclabsustech/DeepSeparator}.
\end{abstract}

\vspace{2pc}

\maketitle
%
%
%
%
%

\section{Introduction}

EEG signals are recorded at the scalp and reflect the electrophysiological activities on the cerebral cortex~\cite{cai2020feature,oberman2005eeg,wolpaw1991eeg,herrmann2005human,liu2017detecting,zhao2019hand}. EEG is important for neuroscience research and clinical applications, such as brain-computer interface (BCI)~\cite{arvaneh2011optimizing,li2011real}, diagnosis of neurological disorders~\cite{von2010finding,zhao2012eeg}. However, EEG recordings contain not only the neural activity, but also a variety of noise and artifacts, including ocular artifacts~\cite{flexer2005using}, myogenic artifacts~\cite{mcmenamin2010validation}, 
cardiac artifacts~\cite{jorge2019investigating, marino2018heart} and non-physiological noises~\cite{lai2018artifacts,jiang2019removal}. These artifacts will greatly impact the results of EEG data analysis, even completely alter our interpretation of results. Thus, it is necessary to develop effective algorithms to reduce the artifacts mixed in the EEG recordings while preserving the neural information as much as possible.

For the EEG signals and artifacts with different spectral profiles, a straightforward way to remove artifacts from EEG signals is to transform the signal from the time domain to the spectral domain using Fourier transform or wavelet transform, and then filter the artifacts-related spectral components. The denoised signal can be reconstructed by inverse Fourier transform or inverse wavelet transform. A variety of filters can be used for EEG denoising, such as Wiener filter \cite{somers2018generic} and Kalman filter \cite{shahabi2012eeg}. 
However, due to the overlap between the artifacts and the EEG spectrum~\cite{allen2000method}, the artifacts cannot be completely removed and the neural information might get lost after filtering.


Other methods aim to transforming the signal from the original space to a new space, so that the signal and noise are separable in this new space, such as adaptive filter~\cite{correa2007artifact}, , Hilbert-Huang Transformation (HHT)~\cite{zhang2009method, gao2015ica,HHT2013clustering}, empirical mode decomposition (EMD)~\cite{wang2015removal, marino2018adaptive}, independent component analysis (ICA)~\cite{nam2002independent,urrestarazu2004independent} and canonical correlation analysis (CCA)~\cite{de2006canonical}. However, these methods mainly use the linear transformation, and they require additional information or heavily rely on prior assumptions. 
Specifically, adaptive filter requires recording artifactual signals as the reference, such as electrooculogram (EOG) signals. The denoising performance may be rather poor if the reference signal is not properly provided. HHT-based artifact removal approach assumes that the artifactual components have distinctive time-frequency features with others. EEG signals are decomposed into Intrinsic Mode Functions (IMF) adaptively. HHT outputs the IMF's instantaneous frequency (IF), which enhance the time-frequency information \cite{zhang2009method, HHT2013clustering}. The IMFs whose IF have large distances from others are selected as noisy components and removed. The threshold of distance is manually designed \cite{HHT2013clustering}.
Both EMD-based and ICA-based approaches decompose the multi-channel EEG signals into multiple modes or components, and then remove the noise-related components according to specific criterion~\cite{huang1998empirical, chen2017use,looney2008ocular}.
Specifically, ICA-based EEG denoising methods rely on two assumptions: i) the neural signal and artifacts are constructed by mutually independent sources; ii) the neural sources and artifactual sources are linearly separable \cite{albera2012ica, hyvarinen2000independent}. 
CCA-based EEG denoising methods are widely used for electromyography (EMG) artifact removal, assuming that the muscle artifacts have low autocorrelation and rare stereotyped topographies \cite{jiang2019removal,lai2018artifacts}. CCA decomposes the EEG signals into several uncorrelated components, and the component with the least autocorrelation are selected as the muscle artifacts to be removed.



Some hybrid methods are recently proposed, such as EEMD-ICA \cite{zeng2015eemd} and EEMD-CCA \cite{chen2018novel}. The combination of traditional methods leads to performance improvements, but still does not solve the limitations of prior assumptions. For example, the selection of two autocorrelation thresholds in EEMD-CCA is determined empirically in different scenarios~\cite{sweeney2012use}.

With the development of deep learning (DL), some classic DL models have also been applied to EEG artifact removal, such as auto-encoder~\cite{yang2018automatic,leite2018deep}, residual convolution neural networks~\cite{sun2020novel,zhang2020eegdenoisenet}, recurrent neural networks~\cite{zhang2021novel,zhang2020eegdenoisenet}, and Generative Adversarial Networks~\cite{sawangjai2021eeganet}. Compared with traditional models, DL models have the following two advantages: 1) \textit{universality}, a uniform architecture can suit diverse artifacts removal tasks without manual designs of prior assumptions on a specific type of artifacts; 2) \textit{higher capacity}, deep learning enforces significant performance improvement. However, the weak interpretability and safety issues of deep learning largely limited its applications to EEG denoising. Thus, interpretable and reliable DL models for EEG denoising are of high interest.

In this paper, we propose an end-to-end DL framework, called DeepSeparator, which learns to remove artifacts from single-channel EEG. 
We employ a novel training strategy to simultaneously train DeepSeparator with EEG, EOG and EMG inputs. The explicit exposure to EOG and EMG artifacts can help DeepSeparator better learn and distinguish the intrinsic characteristics of artifacts mixed in EEG signals.
Our main contributions are summarized as follows:
\begin{itemize}
\item{\textbf{Novel architecture}: DeepSeparator is an end-to-end deep learning framework which does not rely on manually designed prior assumptions and knowledge of artifacts. It can be considered as a nonlinear decomposition and reconstruction of the input, as an extension of linear blind source separation methods. DeepSeparator learns to decompose the clean EEG signal and artifacts in the latent space for single channel EEG, as ICA does for multi-channel EEG denoising.}
\item{\textbf{Strong interpretability}: 
Compared with other deep learning models, the network design of DeepSeparator fosters its interpretability. Specifically, the encoder is responsible for capturing and amplifying the features in the raw EEG, the decomposer for extracting the trend, detecting and suppressing the artifacts in the embedding space, and the decoder for reconstructing the EEG signal and artifact.}
\item{\textbf{High capacity}:  DeepSeparator can deal with various artifacts, such as EOG and EMG. It reliably achieves better performance compared to traditional EEG denoising methods (e.g., adaptive filter, HHT, EEMD-ICA) across multiple SNR levels. The DeepSeparator trained with single-channel, semi-synthetic EEG data can be applied in multi-channel, real EEG data.}

\end{itemize}



\section{Methods}
\label{sec:methods}

Here we present a novel deep learning based architecture for EEG artifacts removal, called DeepSeparator. DeepSeparator is an end-to-end network which learns a non-linear transformation to separate the artifact and the clean neural signal from the raw EEG signal. 
In the following subsections, we demonstrate details on DeepSeparator with respect to model structure, training strategy, datasets for experiments and the validation metrics.

\subsection{Model structure}

\begin{figure*}[h]
\centering
\includegraphics[width=1\textwidth]{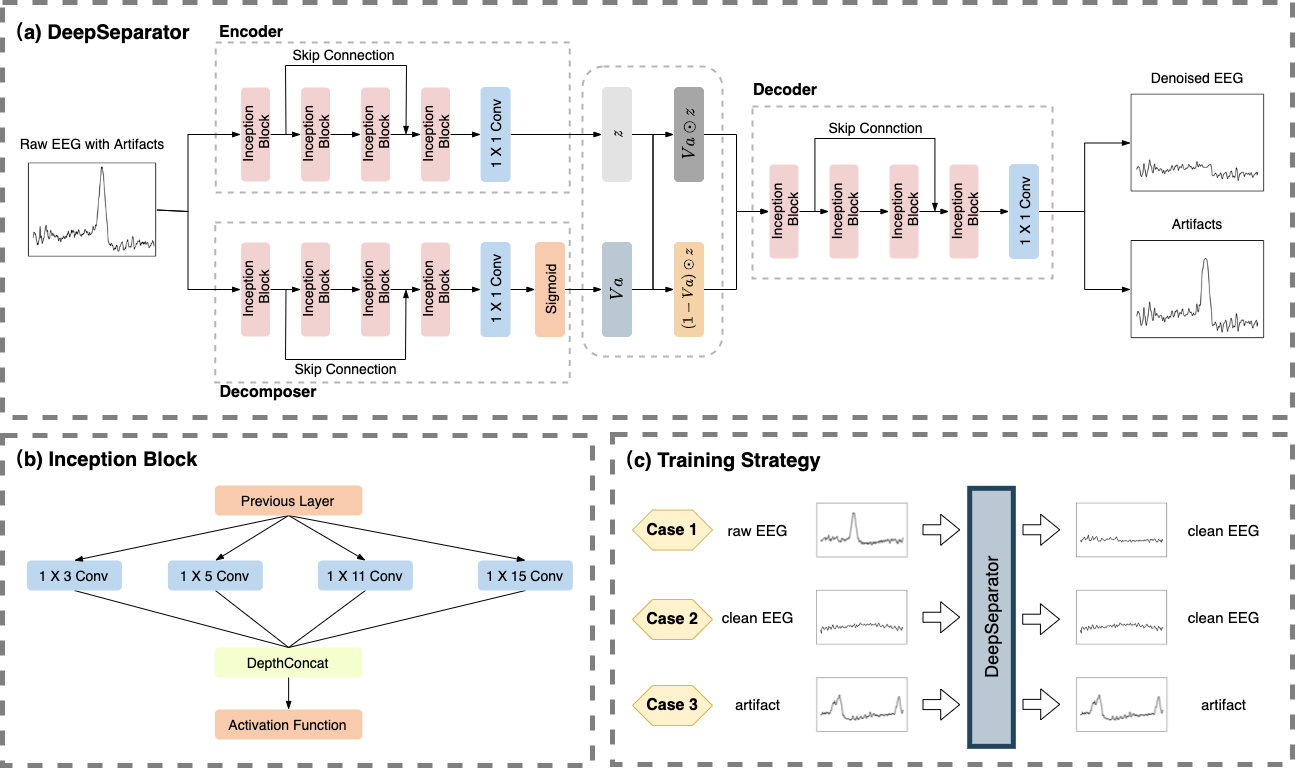}
\caption{Network structure for DeepSeparator. (a) DeepSeparator. Stacked Inception block constitutes the encoder, decomposer and decoder of DeepSeparator. Encoder extracts and amplifies the features of the original signal. Decomposer identifies and suppresses the artifact mixed in the EEG signal. Decoder reconstructs the processed signal to output denoised EEG. (b) Inception block. It consists of four kernel with different size to capture feature of different scale. Result of convolution operation are concatenated in the dimension of channel. (c) The training strategy. There are three different input-to-output pairs for training: rawEEG-to-cleanEEG (case 1), cleanEEG-to-cleanEEG (case 2) and artifact-to-artifact (case 3). Both EEG and artifact participate in the training, facilitating the learning and generalizability. }
\label{fig:framework}
\end{figure*}

We design the structure of DeepSeparator with an encoder, a decomposer and a decoder as depicted in Fig. \ref{fig:framework}(a). 
Specifically, the \textit{encoder} transforms the input to an embedding vector $z$. In the embedding space, each element of the embedding vector is assumed to associate with either artifact or signal component. 
The \textit{decomposer} learns an attenuation vector $V_{atte}$ in which the value of each element is limited to the interval between 0 and 1. Element-wise multiplication between embedding vector and attenuation vector (i.e., $v_{atte} \odot z$ or $(1 - v_{atte}) \odot z$) decides what information should be thrown away or kept, which is similar to the forget gate in LSTM and GRU. The \textit{decoder} reconstructs either the signal or the artifact according to the hyperparameter $v_{indicator}$ we defined. The procedures are formulated as follows:
\begin{equation} 
\begin{aligned}
&\text{Encoder}: \ z = f_\theta(x), \\
&\text{Decomposer}: \ v_{atte} = f_\psi(x), \\
&\text{Attenuated Embedding}: \  \tilde{z} = |v_{indicator} - v_{atte} | \odot z,\\
&\text{Decoder}: \ \hat{y} = f_\phi(\tilde{z}) ,
\end{aligned}
\end{equation}
where  $f_\theta$ is the encoder; $f_\psi$ is the decomposer; $f_\phi$ is the decoder; $ z $ denotes the embedding vector;  $ v_{atte} $ denotes the attenuation vector; $ v_{indicator} $ denotes the indicator vector indicating the model to output signal or artifact; $ \odot $ denotes the element-wise product; $ \tilde{z} $ denotes the attenuated embedding vector; $ \hat{y}  $ denotes the extracted signal or artifact. 

Here we use the Inception Block \cite{szegedy2015going} as the basic component of DeepSeparator as shown in Fig.~\ref{fig:framework}(b). The Inception Block stacks convolution kernels of different sizes in the same layer. We use the size of $1\times3$, $1	\times 5$, $1\times 11$, $1\times 15$. EEG data often contains artifacts and noise at different frequencies, such as EOG and EMG artifact. Kernels with multiple sizes can capture feature at various scales to help the model to better learn the characteristics of EEG signals and different artifacts and noise, so as to achieve better artifact removal performance. By stacking Inception blocks, we construct the encoder, decomposer and decoder. Besides, we do not introduce any fully connected layer in this model, and keeping the shape of input and output of each layer unchanged by padding, so DeepSeparator can process input of any length. To be noted, besides the Inception Block, other arbitrary neural models might be also valid for the encoder, decomposer and decoder of DeepSeparator.

\subsection{Training and testing}
\label{sec:training}
\subsubsection{Training strategy} We propose a new training strategy to facilitate the learning of the characteristic features from neural signals and artifacts (Fig.~\ref{fig:framework}(c)). Specifically, we designed three training cases with different input-to-output pairs: RawEEG-to-cleanEEG (case 1), cleanEEG-to-cleanEEG (case 2) and artifact-to-artifact (case 3). The indicator vector $v_{indicator}$ represents a specific training case. When the output is clean EEG, all elements of the indicator vector are set to 0 (i.e. cases 1 \& 2), otherwise all elements are set to 1 (i.e. case 3).
DeepSeparator is trained in a supervised manner, with the Mean Square Error (MSE) between the outputs and the expected reconstructed clean EEG or artifacts as loss function. 
A major benefit of our training strategy is that both clean EEG and pure artifacts participate in the training of the model, so that the model can directly learn the distinguishable characteristics of clean EEG and artifacts. 
Compared with other DL models trained only in RawEEG-to-cleanEEG case~\cite{yang2018automatic,leite2018deep,sun2020novel,zhang2020eegdenoisenet,zhang2021novel}, our training strategy will facilitate learning and generalizability. 
\re{To train the network, the Adam optimization algorithm is used, and the size of the minibatch and the initial learning rate are set to 2048 and 0.001.}


\subsubsection{Testing} After training, DeepSeparator can be applied to remove artifacts from single-channel EEG. The indicator vector can be a hyperparameter to control DeepSeparator to extract the denoised EEG signal or the artifact in the test: 0 to output the denoised EEG, 1 to output the artifacts. 
Moreover, multi-channel EEG signals can input DeepSeparator channel by channel. The effects of EEG denoising may influence the downstream EEG tasks, such as event-related-potential (ERP) analysis and EEG source localization, which requires further testing.

\subsection{Datasets}

We use a semi-synthetic EEG dataset (i.e., EEGdenoiseNet \cite{zhang2020eegdenoisenet}) and a real EEG dataset (i.e., MNE-SAMPLE-DATA \cite{ds000248:1.2.1}) in this paper. 
EEGdenoiseNet provides noisy EEG and corresponding clean EEG for model training and quantitative evaluation of performance. MNE-SAMPLE-DATA provides EEG data collected in experiments without ground truth result. It can be used to judge the generalization ability of different models in real scenarios.

\subsubsection{Semi-synthetic EEG} EEGdenoiseNet~\cite{zhang2020eegdenoisenet} contains 4514 clean EEG segments, 3400 ocular artifact segments and 5598 muscular artifact segments. Each EEG segments have 512 samples (2 seconds). The contaminated EEG can be generated by linearly mixing the clean EEG segments with EOG or EMG artifact segments according to equation: 

\begin{equation}  
y = x + \lambda \cdot n,
\end{equation}
where y denotes the mixed signal of EEG and artifacts; x denotes the clean EEG signal as the ground truth; n denotes (ocular or myogenic) artifacts; $\lambda$ is a hyperparameter to control the signal-to-noise ratio (SNR) in the contaminated EEG signal y. Specifically, the SNR of the contaminated segment can be adjusted by changing the parameter $\lambda$ according to Eq.~\eqref{eq:SNR}:

\begin{equation}  \label{eq:SNR}
SNR = 10 \times log\frac{RMS(x)}{RMS(\lambda \cdot n)},
\end{equation}
The root mean square (RMS) is defined as:
\begin{equation} \label{eq:RMS}
RMS(x) = \sqrt{\frac{1}{N} \sum_{i=1}^N x_i^2},
\end{equation}
where N denotes the number of temporal samples in the segment $x$, and $x_i$ denotes the $i^{th}$ sample of segment x.

\re{Under a specific SNR, one clean EEG segment can be linearly mixed with an arbitrary ocular artifact segment or a muscular artifact segment, generating up to 3400+5598 types of contaminated EEG.
In total, we synthesize 25,000 noisy EEG segments containing EOG artifacts and 25,000 noisy EEG segments containing EMG artifacts, totally 50,000 noisy EEG segments. 
We used 40,000 noisy EEG segments (20,000 EEG with EOG plus 20,000 EEG with EMG) for training and 10,000 noisy EEG segments (5,000 EEG with EOG plus 5,000 EEG with EMG) for testing. Because the training of DeepSeparator requires artifact data, 3400 ocular artifact segments and 5598 muscular artifact segments in EEGdenoiseNet are introduced for training.}

\subsubsection{Real EEG} 
We use a real EEG dataset, MNE-SAMPLE-DATA~\cite{ds000248:1.2.1}, to test the generalizability of the trained DeepSeperator. MNE-SAMPLE-DATA contains 59-channel EEG signals recorded at task by the Neuromag Vectorview system at MGH/HMS/MIT Athinoula A. Martinos Center Biomedical Imaging. 
In the task, checkerboard patterns are presented into the left and right visual field, interspersed by tones to the left or right ear. The interval between the stimuli is 750 ms. 
We apply the trained DeepSeparator, as well as ICA, to this real EEG dataset after filtering. Then, we epoch the EEG signals 100ms prior auditory events and 400ms post stimulus. We average across trials to obtain ERP signals.
The neural sources of ERP components are localized using MNE-Python toolbox with BEM head model and MNE source localization method.


\subsection{Validation metrics}

To quantitatively evaluate the performance of models, three metrics are used on the test dataset, including RRMSE in the temporal domain, RRMSE in the spectral domain and the temporal correlation coefficient ($CC$).

\begin{equation}  
RRMSE_t = \frac{RMS(\hat{y}-y)}{RMS(y)},
\end{equation}

\begin{equation} 
RRMSE_s = \frac{RMS(PSD(\hat{y})-PSD(y))}{RMS(PSD(y))},
\end{equation}

\begin{equation} 
CC =\frac{Cov(\hat{y},y)}{\sqrt{Var(\hat{y})Var(y)}},
\end{equation}
where $RMS$ is defined as Eq.~\eqref{eq:RMS}; $PSD$ denotes to the power spectral density of the input data; $Var$ and $Cov$ denote the variance and covariance, respectively.

\section{Experiments and Results}
\label{sec:exp}

\subsection{Results of semi-synthetic EEG data}

We first test the denoising performance of DeepSeparator with a semi-synthetic EEG dataset where we have the ground truth. The results are compared with adaptive filter, HHT and EEMD-ICA.

\subsubsection{EOG artifact elimination}

\re{We quantify the mean performance of EOG artifact removal over 10,000 test samples with three metrics, namely temporal RRMSE, spectral RRMSE and CC (see Table \ref{tab:EOG_value}). }
DeepSeparator obtains the lowest temporal RRMSE, lowest spectral RRMSE and highest CC with lower standard variance, suggesting that DeepSeparator outperforms the traditional denoising methods in EOG artifact removal.

\begin{table}[htbp]\scriptsize
\renewcommand\arraystretch{1.5}
\caption{Performance comparisons of EOG artifact removal. \re{The mean $\pm$ std of 3 metrics from test samples are presented.}}
\begin{center}
\begin{tabular}{c|ccc}
\hline
Model & Temporal RRMSE & Spectral RRMSE  & CC \\
\hline
Adaptive Filter & \re{1.213} $\pm$ \re{0.697} & \re{1.186} $\pm$ \re{0.668} & \re{0.562} $\pm$ \re{0.173}   \\
HHT             & \re{2.375} $\pm$ \re{1.569} & \re{2.346} $\pm$ \re{1.630} & \re{0.418} $\pm$ \re{0.221}  \\
EEMD-ICA        & \re{1.330} $\pm$ \re{0.675} & \re{1.334} $\pm$ \re{0.785} & \re{0.289} $\pm$ \re{0.207}  \\
DeepSeparator   & \re{0.705} $\pm$ \re{0.249} & \re{0.747} $\pm$ \re{0.302} & \re{0.769} $\pm$ \re{0.142}  \\
\hline
\end{tabular}
\label{tab:EOG_value}
\end{center}
\end{table}


\begin{figure}[htb]
\centering
\includegraphics[width=0.8\textwidth]{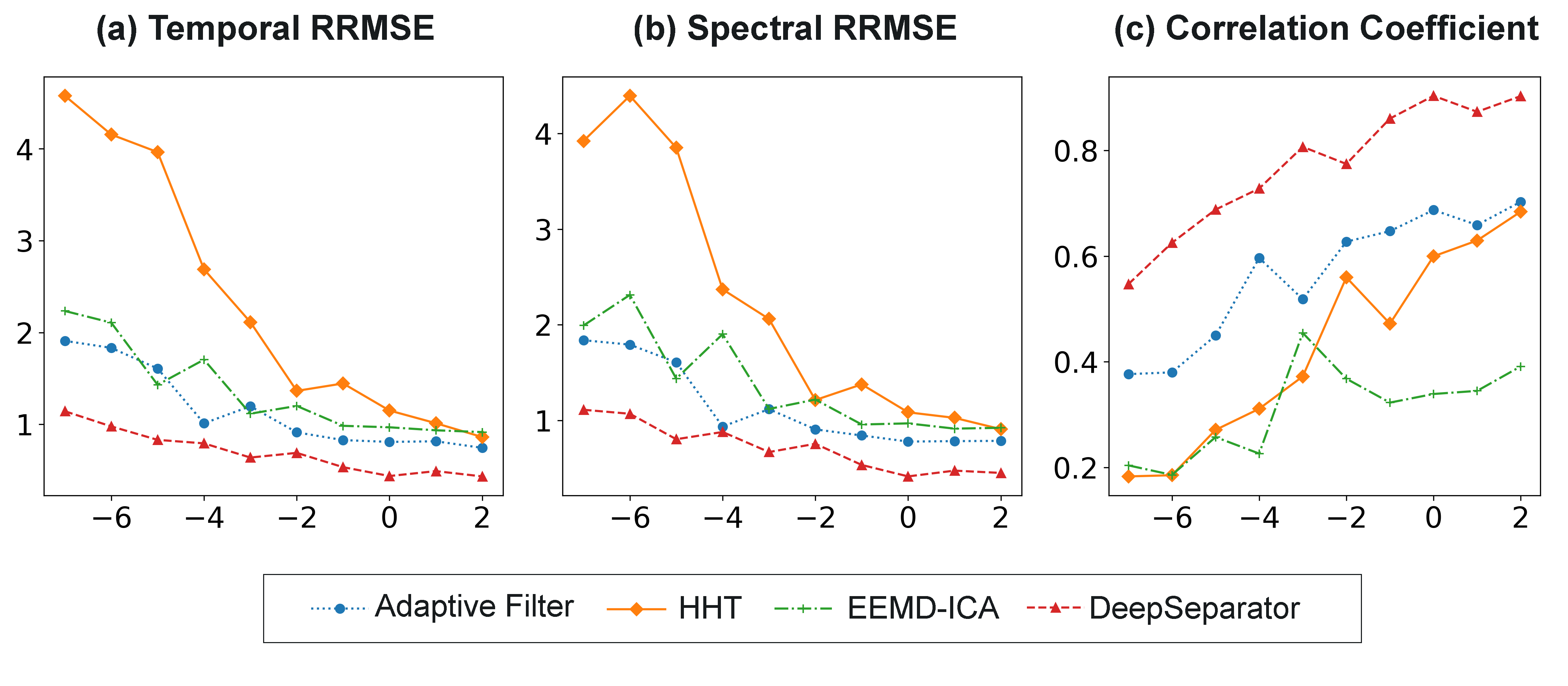}
\caption{EOG artifact removal at multiple SNR levels. The SNR is ranging from -7 to 2. DeepSeparator reliably obtains lower RRMSE temporal (left), RRMSE spectral (middle) and CC (right), suggesting that DeepSeparator outperforms the adaptive filter, HHT and EEMD-ICA methods under different SNRs.}
\label{fig:snr_eog}
\end{figure}

To further investigate the robustness of our model in noisy EEG, we test the performance at multiple SNR levels (-7 to 2 dB in Eq.~\eqref{eq:SNR}). The quantitative results are shown in Fig.~\ref{fig:snr_eog}. Generally, the performance of all four methods becomes worse with increasing noise (i.e. decreasing SNR). But DeepSeparator reliably outperforms other methods across SNR levels with lower RRMSE and higher CC. The results suggest DeepSeparator is robust even when there are large EOG artifacts in the input.

\begin{figure}[htb]
\centering
\includegraphics[width=0.65\textwidth]{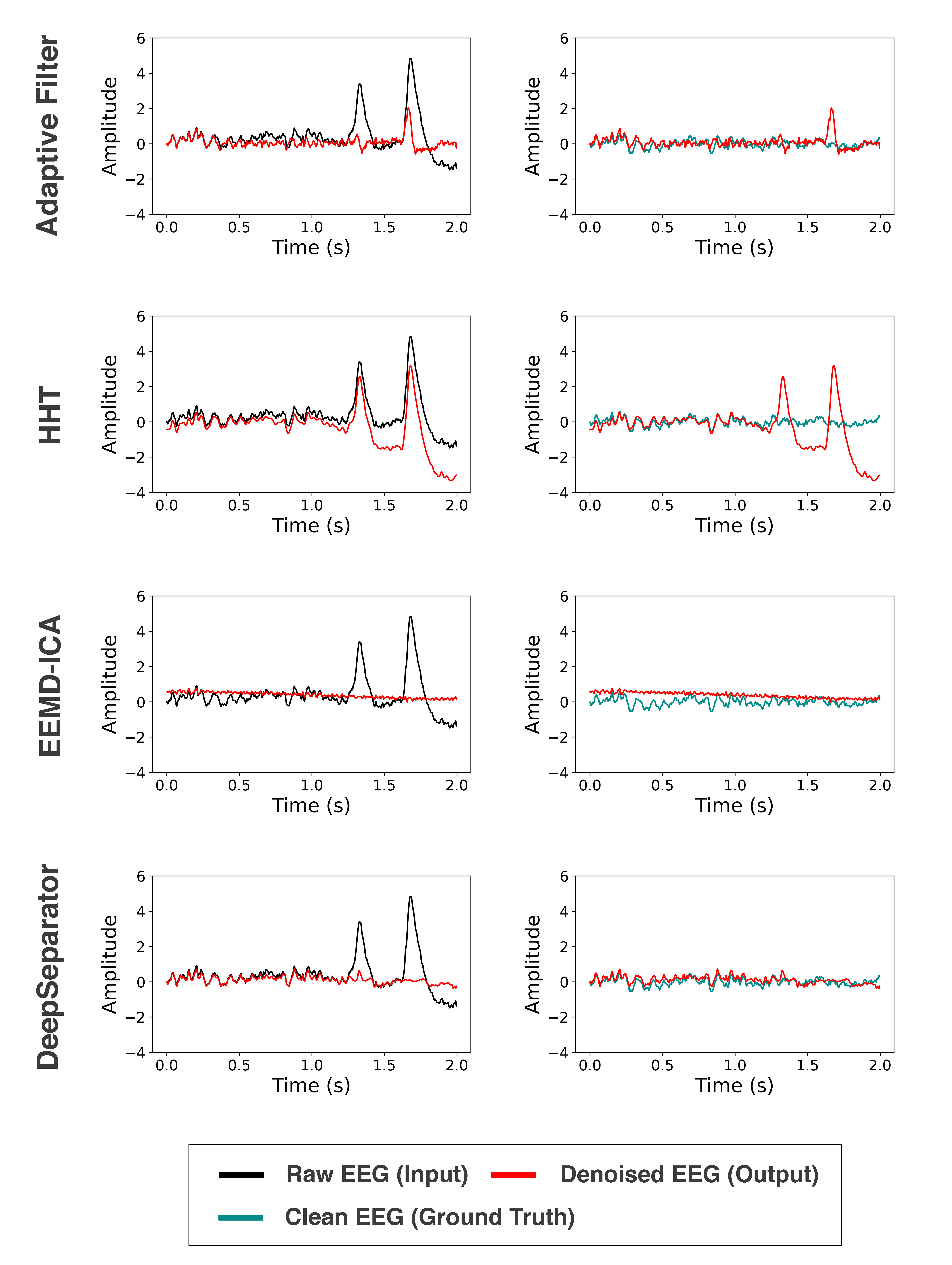}
\caption{An example of EOG artifact removal for a test EEG data. From top to bottom, we present the results using Adaptive Filter, HHT, EEMD-ICA and DeepSeparator. The raw EEG (input), the denoised EEG (output) and the clean EEG (ground truth) are shown with the black line, red line, and green line, respectively. There are two EOG peaks in the raw data, and DeepSeparator can largely suppress them.}
\label{fig:EOG_plot}
\end{figure}

To visualize the denoising performance, we present a sample of EEG time courses of raw EEG (model input), clean EEG (ground truth) and denoised EEG (model output) using adaptive filter, HHT, EEMD-ICA and DeepSeparator (see Fig.~\ref{fig:EOG_plot}). It shows that DeepSeparator can effectively suppresses EOG artifact and retain the clean EEG signal as much as possible. This phenomenon is consistent with the quantification of metrics in Table~\ref{tab:EOG_value}. In contrast, adaptive filter and HHT cannot effectively suppress the EOG peaks. Although EEMD-ICA can suppress the EOG peak, it also loses a lot of information, resulting in denoised EEG close to a straight line. Besides, the result of EEMD-ICA are unstable, with significant changes of repeated experiment for the same data. Fig.~\ref{fig:EOG_plot} indicates DeepSeparator can mostly recover the ground truth EEG signal.

\begin{figure}[htb]
\centering
\includegraphics[width=0.7\textwidth]{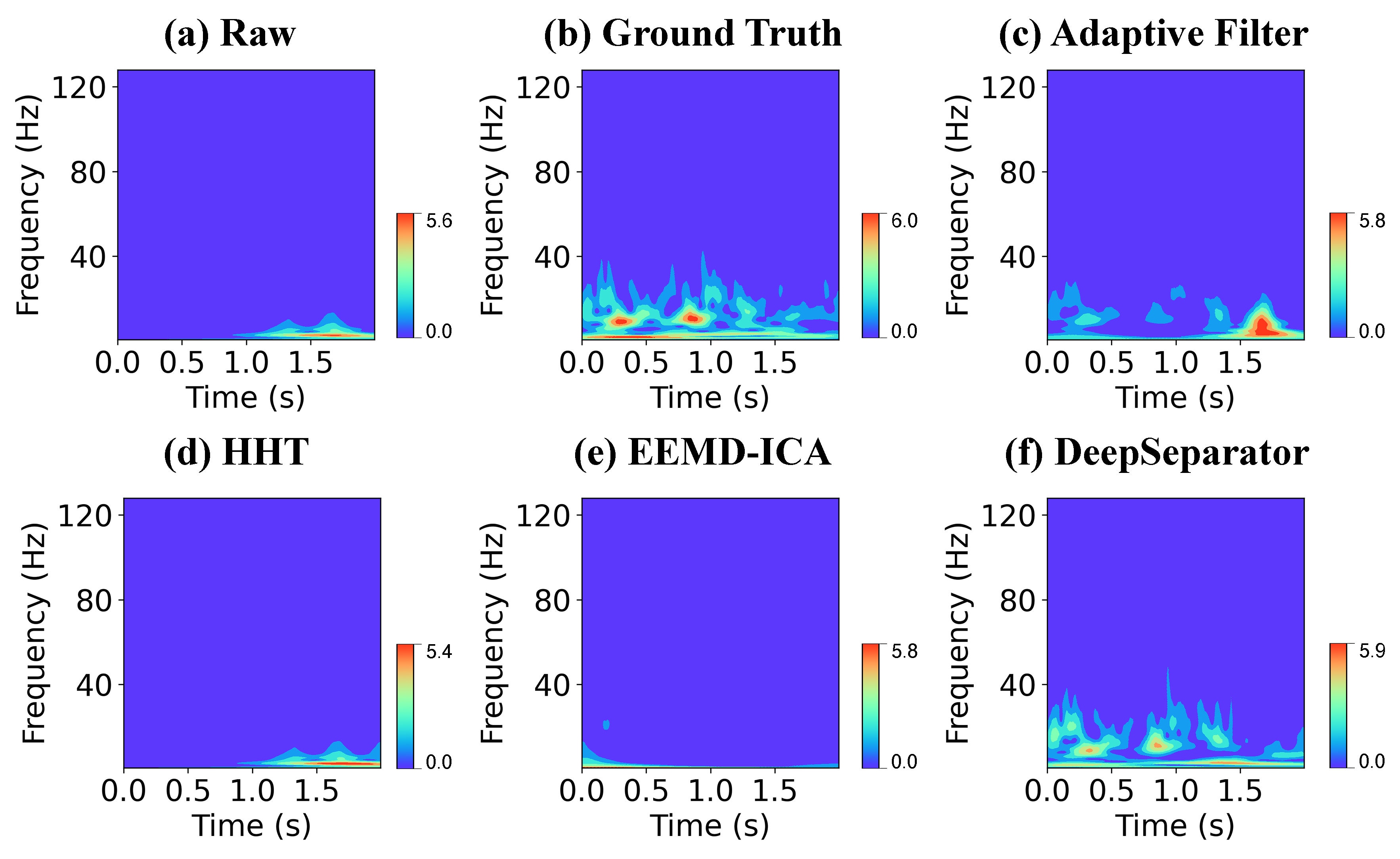}
\caption{Time-frequency analysis to examine EOG artifact removal. We plot the spectrogram of the raw EEG (a), the clean EEG as ground truth (b), and the denoised EEG as output of adaptive filter (c), HTT (d), EMDD-ICA (e), DeepSeparator (f), respectively. }
\label{fig:EOG_freq}
\end{figure}

We then conduct the time-frequency analysis to examine the performance of removing EOG artifact. Fig.~\ref{fig:EOG_freq}(a)-(f) present the spectrograms from the raw EEG signal, the ground-truth EEG, and the denoised EEG signal using Adaptive Filter, HHT, EEMD-ICA and DeepSeparator respectively. It is obvious that the spectrogram of denoised EEG by DeepSeparator is the most resemblance to the ground-truth EEG. 
In contrast, the adaptive Filter, HHT and EEMD-ICA distort the spectrogram. These results suggest that DeepSeparator capture the time-frequency information of EEG data and can reconstruct the spectrogram of the clean EEG data.



\subsubsection{EMG artifact elimination}

The quantification of performance for denoising EMG artifact is shown in Table \ref{tab:EMG_value}. The results show that our proposed model is much better than other competing models in terms of all three metrics.

\begin{table}[htbp]\scriptsize
\renewcommand\arraystretch{1.5}
\caption{Performance comparisons of EMG artifact removal. \re{The mean $\pm$ std of 3 metrics from test samples are presented.}}
\begin{center}
\begin{tabular}{c|ccc}
\hline
Model & Temporal RRMSE & Spectral RRMSE  & CC \\
\hline
Adaptive Filter & \re{2.348} $\pm$ \re{1.340} & \re{2.333} $\pm$ \re{1.341} & \re{0.346} $\pm$ \re{0.166}  \\
HHT             & \re{1.718} $\pm$ \re{1.161} & \re{1.705} $\pm$ \re{1.151} & \re{0.527} $\pm$ \re{0.217} \\
EEMD-ICA        & \re{1.518} $\pm$ \re{0.965} & \re{1.512} $\pm$ \re{0.990} & \re{0.217} $\pm$ \re{0.218}  \\
DeepSeparator   & \re{0.712} $\pm$ \re{0.275} & \re{0.717} $\pm$ \re{0.295} & \re{0.734} $\pm$ \re{0.175}  \\

\hline
\end{tabular}
\label{tab:EMG_value}
\end{center}
\end{table}

\begin{figure}[htb]
\centering
\includegraphics[width=0.8\textwidth]{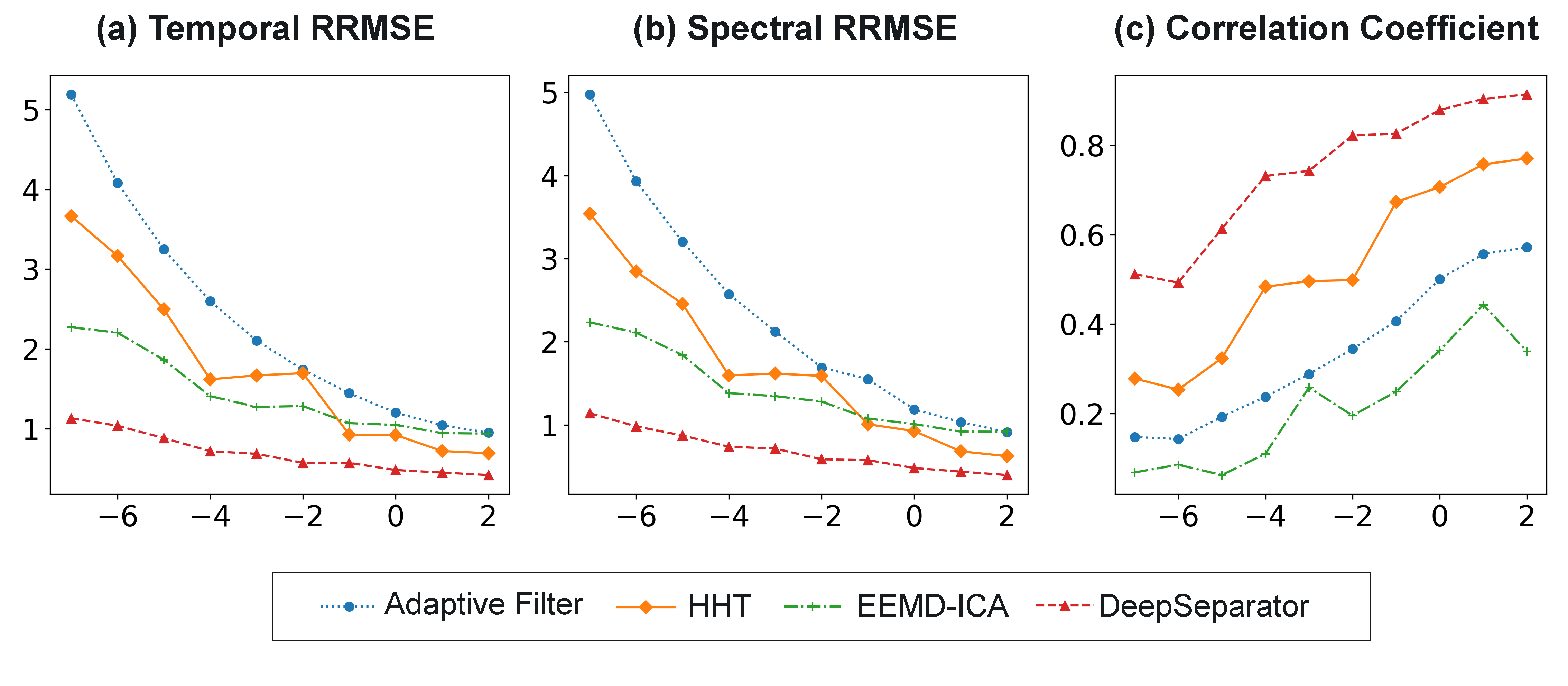}
\caption{EMG artifact removal performance of different models on different SNR:  DeepSeparator achieve the best overall performance and is similar to EOG artifact removal.}
\label{fig:snr_emg}
\end{figure}

We quantitatively evaluated the EMG artifact removal performance under different noise levels. The results in Figure \ref{fig:snr_emg} are a little different from the result of EOG, while DeepSeparator is still the best model. 

\begin{figure}[htb]
\centering
\includegraphics[width=0.65\textwidth]{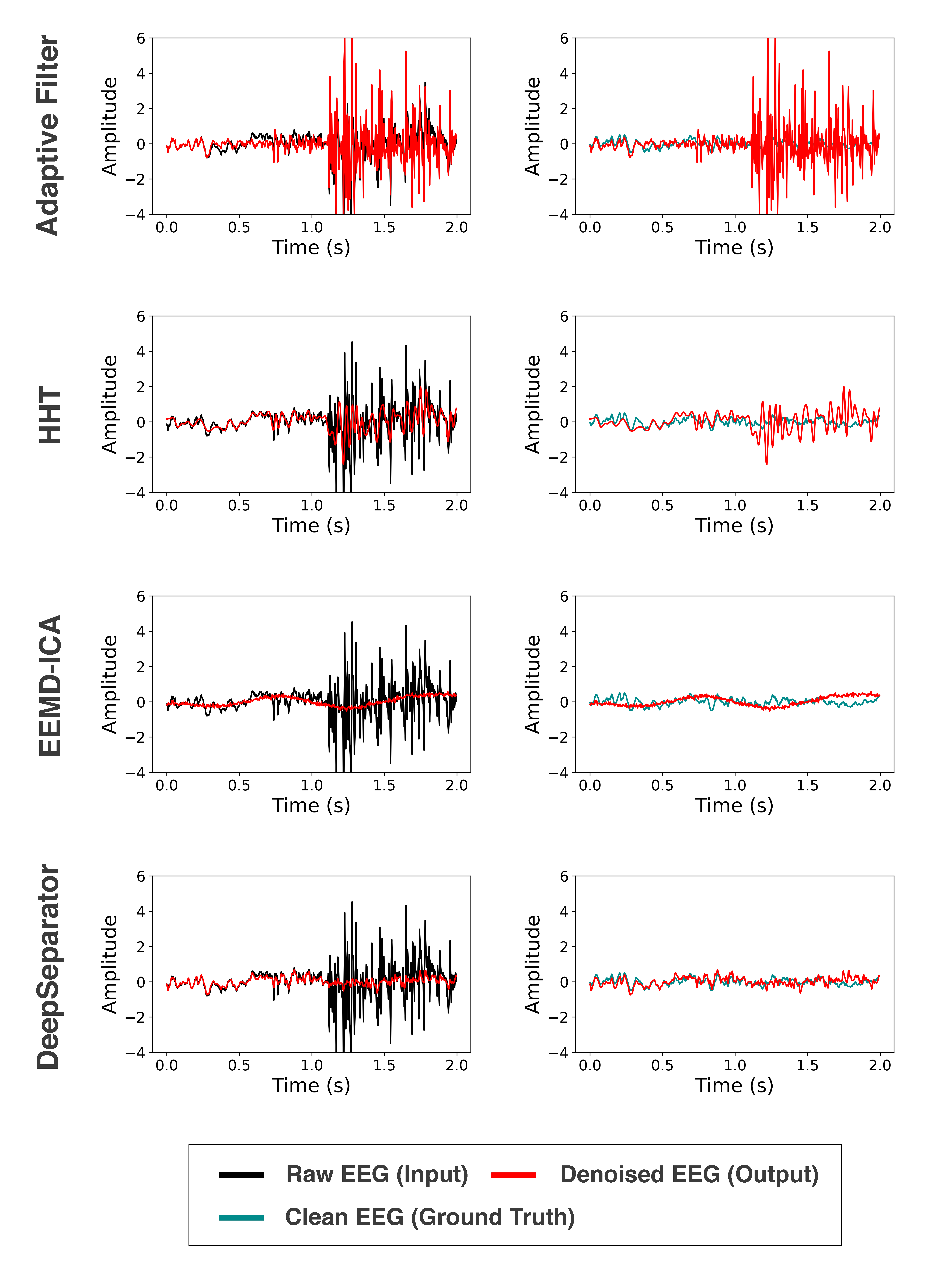}
\caption{An example of EMG artifact removal for a test EEG data. From top to bottom, we present the results using Adaptive Filter, HHT, EEMD-ICA and DeepSeparator. The raw EEG (input), the denoised EEG (output) and the clean EEG (ground truth) are shown with the black line, red line, and green line, respectively. DeepSeparator can largely suppress the EMG artifacts at 1-2 second and mostly recover the clean EEG.}
\label{fig:EMG_plot}
\end{figure}

We illustrate the time course of a test data in Fig.~\ref{fig:EMG_plot}. The results demonstrate that DeepSeparator can better suppress the EMG artifact at 1-2 seconds, compare to other competing methods. Specifically, the adaptive filter and HHT cannot remove the existing EMG artifacts, while EEMD-ICA brings additional slow oscillations in the denoised EEG. DeepSeparator can almost completely eliminate the EMG and recover the clean EEG.

\begin{figure}[htb]
\centering
\includegraphics[width=0.7\textwidth]{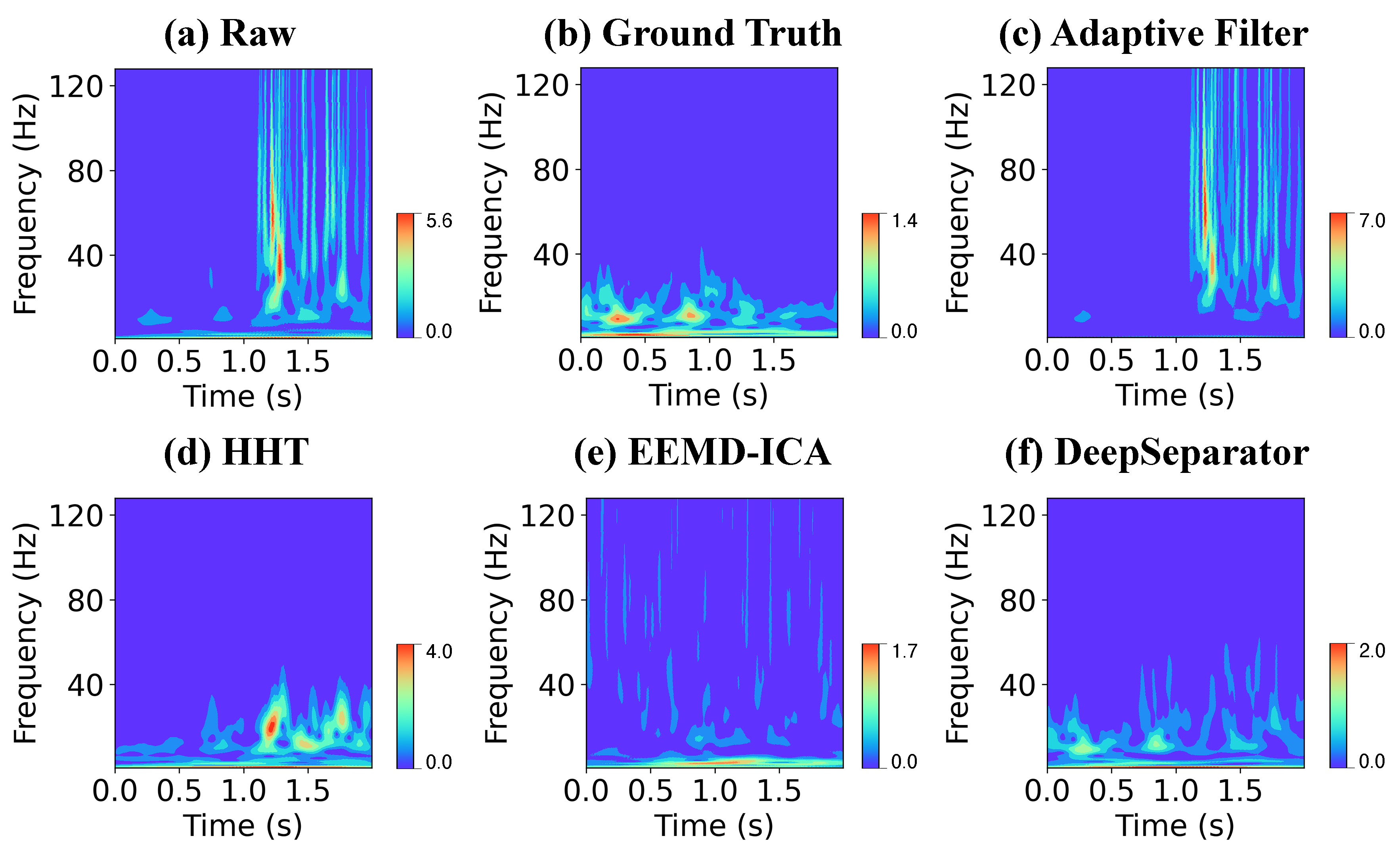}
\caption{Comparison of spectrograms: (a) the raw EEG, (b) the ground truth clean EEG, and the EEG denoised by (c) Adaptive Filter, (d) HTT, (e) EMDD-ICA, (f) DeepSeparator. Raw spectrogram contain high frequency information from 1s to 2s, Adaptive filter and EEMD-ICA cannot filter them effectively. HHT cannot restore the low frequency information from 0s to 1s. DeepSeparator can achieve a better performance while the result is still not close to the ground truth result. Compared with EOG artifact, EMG artifact is more difficult to remove and models still need to be improved.}
\label{fig:EMG_freq}
\end{figure}

Fig. \ref{fig:EMG_freq}(a)-(f) present the result of time-frequency analysis to examine the EMG artifact removal performance, with spectrograms of raw EEG signal, the ground truth EEG, and the denoised EEG signal using Adaptive Filter, HHT, EEMD-ICA and DeepSeparator respectively. Spectrogram of denoised EEG by DeepSeparator is the most resemblance to the ground-truth EEG, while high-frequency artifact from 1s to 2s is still not completely removed. Adaptive filter, HHT and EEMD-ICA cannot suppress the high-frequency artifact, and EEMD-ICA even introduce more noise in the 
period between 0s and 1s. The results suggest EMG artifact is more complicated to be removed and DeepSeparator can still achieve a better performance compared with other models.

\subsubsection{Separation of signal and artifact}

In addition to the satisfactory artifact removal performance, DeepSeparator can extract the signal and artifact from raw input. 
To explore whether the signal and artifact extracted by DeepSeparator are valid and provide more information, we selected some data for visualization and analysis.


\begin{figure*}[htb]
\centering
\includegraphics[width=0.95\textwidth]{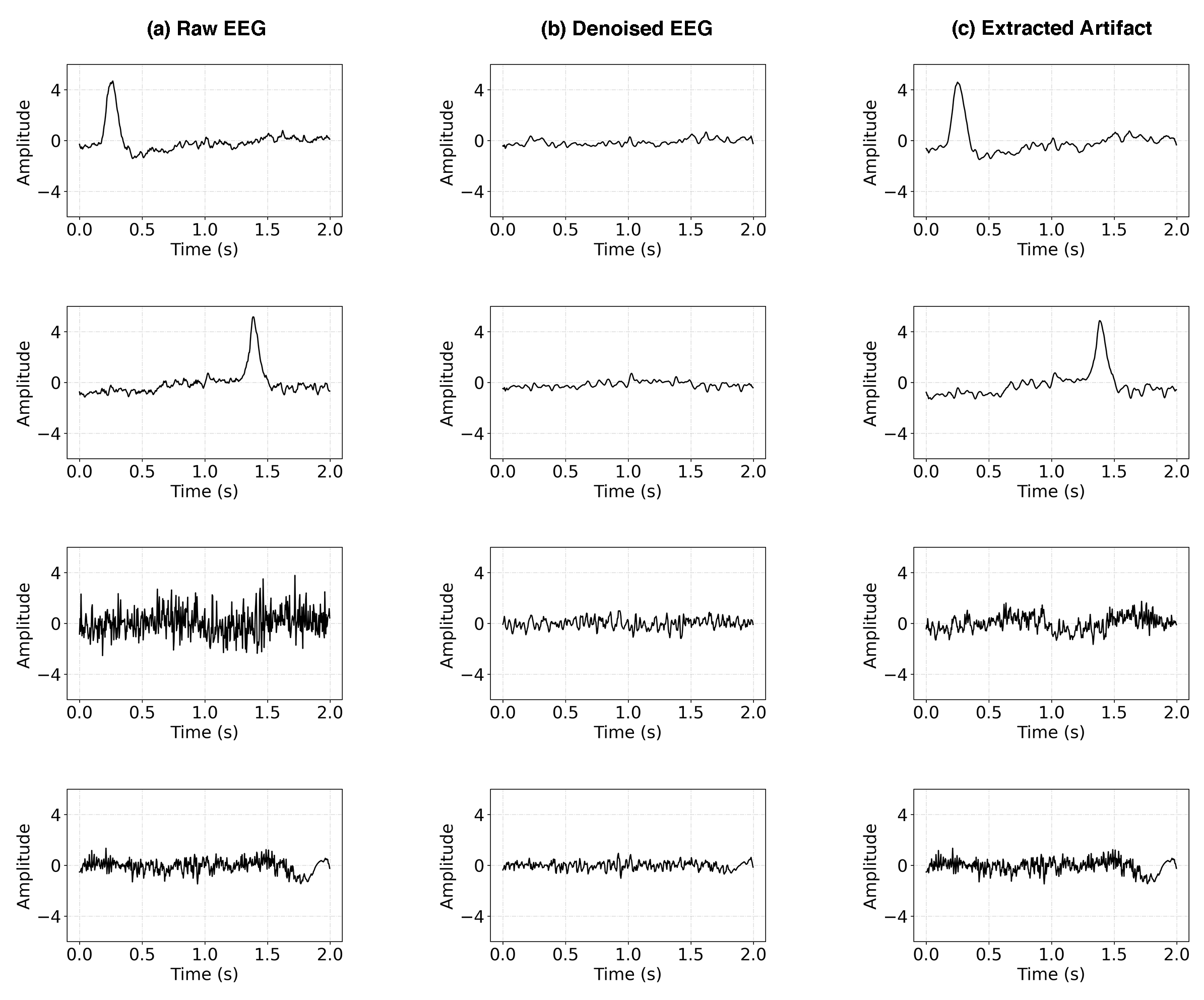}
\caption{Examples of raw EEG (a), denoised EEG (b) and extracted artifact (c) by DeepSeparator. From top to bottom, two EEG with EOG artifact and two EEG with EMG artifact are shown in the figure. 
As for EEG with EOG artifact, EOG artifact is effectively suppressed in the extracted signal, which is retained in the extracted artifact. Beyond the area with EOG artifact, signal of extracted signal and artifact are similar to each other. Both neural signals and artifacts are transmitted in the same medium, and both reflect the activity of the brain over a period of time, so the similarity between denoised EEG and extracted artifact is reasonable
As for EEG with EMG artifact, the extracted signal contain more low frequency information while the extracted artifact contain more high frequency information.}
\label{fig:separation}
\end{figure*}

Fig.~\ref{fig:separation} shows four samples of raw EEG data, extracted signals and artifacts by DeepSeparator. 
For EOG artifact removal, DeepSeparator can effectively suppresses EOG artifact in the extracted signal while the EOG peaks are retained in the extracted artifact.
For EMG artifact removal, extracted signal contains more low frequency information while extracted artifact contains more high frequency information. 
These results demonstrated that there is a significant difference between the extracted signal and artifact, and we can further judge the type of artifact by these difference.

\begin{figure*}[htb]
\centering
\includegraphics[width=1\textwidth]{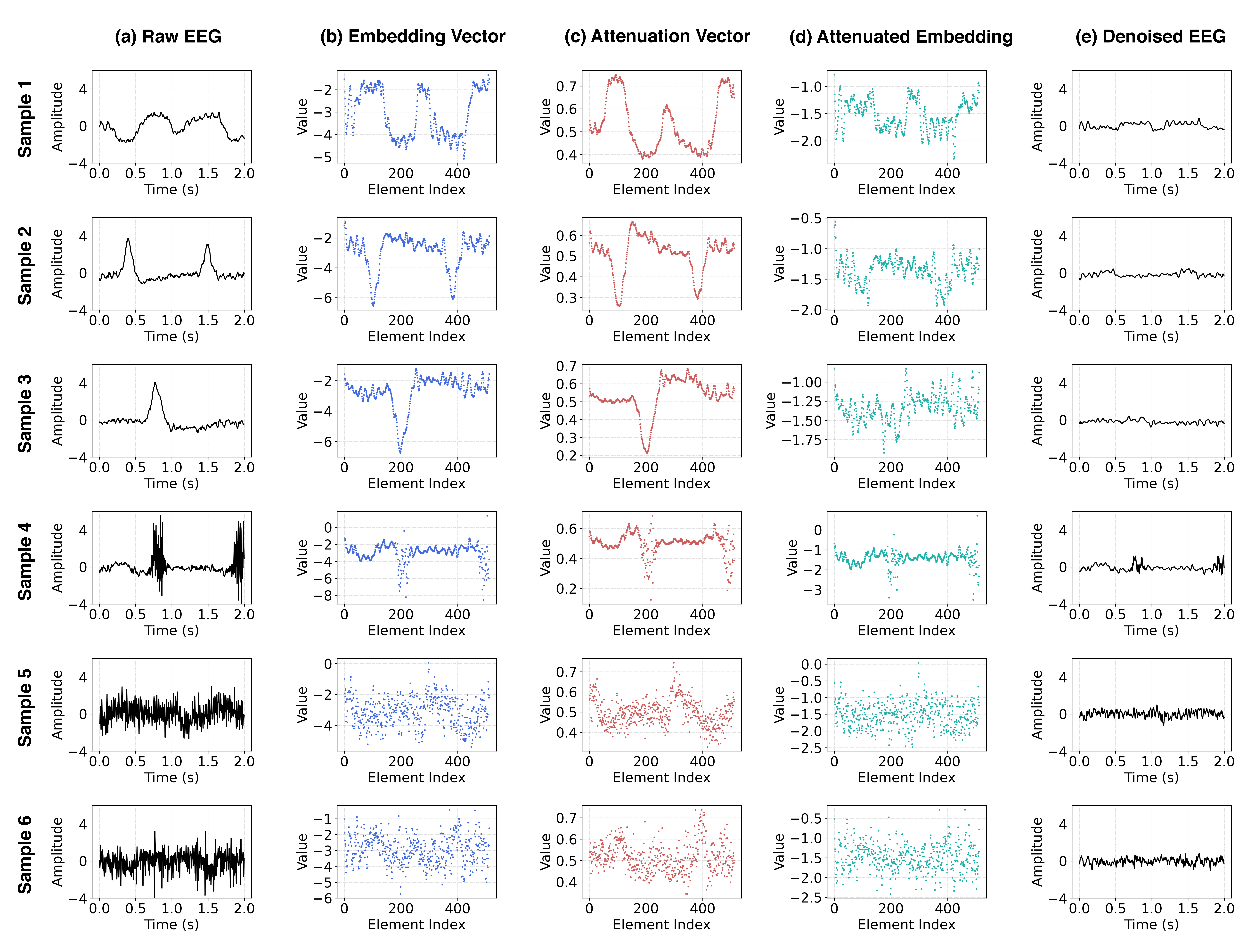}
\caption{Visualization: (a) the raw EEG $x$, (b) embedding vector $z$, (c) attenuation vector $v_{atte}$,  (d) attenuated embedding vector $\tilde{z}$,  (e) denoised EEG $\hat{y}$. 
Only the vector related to artifact removal are shown in figure. From sample 1, sample 2 and sample 3, the fluctuation of raw EEG is amplified in the embedding vector, attenuation vector detect the location of artifact and try to attenuate the artifact, so that the attenuated embedding vector has less artifact compared with embedding vector. At last, the denoised EEG signal is reconstructed by decoder. From sample 4, sample 5 and sample 6, EMG artifact removal mechanism is more complex.}
\label{fig:visulization}
\end{figure*}

Similar to other deep learning models, there is no clear understanding of why DeepSeparator perform so well, and how it might be improved. We explore both issues by visualizing the intermediate feature layers, i.e. embedding vector, attenuation vector and attenuated embedding vector in Fig.~\ref{fig:visulization}. 
From top to bottom, there are three EEG with EOG artifact and three EEG with EMG artifact in the figure. 
As for the EEG with EOG artifact (i.e. sample 1, sample 2 and sample 3), the shape of embedding vector is close to the inverted original EEG, and the fluctuation is more significant. The visualization of attenuation vector shows that the element value of the area artifact introduced is closer to 0. Hence, by element-wise product, the EOG peaks are effectively suppressed in the attenuated embedding vector. The results also revealed that the temporal feature is preserved in the intermediate feature layers of the model.
As for the EEG with EMG artifact (i.e. sample 4, sample 5 and sample 6), EMG seriously disturb the EEG signal. 
Sample 4 is partially disturbed. The visualization of attenuation vector shows that model tends to assign a lower value to suppress mutation of amplitude caused by EMG artifact.
Sample 5 and sample 6 are completely disturbed by EMG artifact, but the element of their attenuation vector are distributed around 0.5, and some element approach 1 when there is a trend in the corresponding area in raw input. We guess that on the one hand, DeepSeparator tries to control the amplitude of signal within a reasonable scale, while at the same time extracting the trend in the data as much as possible by assigning a larger value in attenuation vector.


In general, the visualization of EOG and EMG artifact removal reveals that under the framework of DeepSeparator, encoder tries to amplify the feature of signal, decomposer tries to detect and suppress the artifact and keep the signal amplitude within a reasonable scale by multiplying smaller value, as well as extracting the trend.

\subsection{Results of real EEG data}

Here we transfer the trained DeepSeparator to the real multi-channel EEG data. 
We compared our proposed model with ICA, a widely used method for multi-channel EEG denoising. 
The EEG data is from MNE-SAMPLE-DATA, a 59-channel EEG data recorded during a series of experiments and we focus on the left auditory stimulation.
Specifically, we band-pass filter the raw EEG signals between 0.5 to 75 HZ and notched at 50Hz (i.e. the powerline frequency), followed by denoising using ICA and DeepSeparator respectively. 
\re{On the one hand, the filtered EEG signals are decomposed into 59 independent components by ICA. With ICLabel ~\cite{pion2019iclabel} and check by expert, we select 5 component (IC1, IC6, IC11, IC14 and IC45 in Fig.~\ref{fig:EEG_plot} (b)) as artifact components and exclude them. 
On the other hand, the filtered EEG signals are input to the trained DeepSeparator channel by channel, and the extracted artifacts and the denoised EEG signals are output by the network.}


\subsubsection{EEG time course}
\begin{figure*}[htb]
\centering
\includegraphics[width=0.95\textwidth]{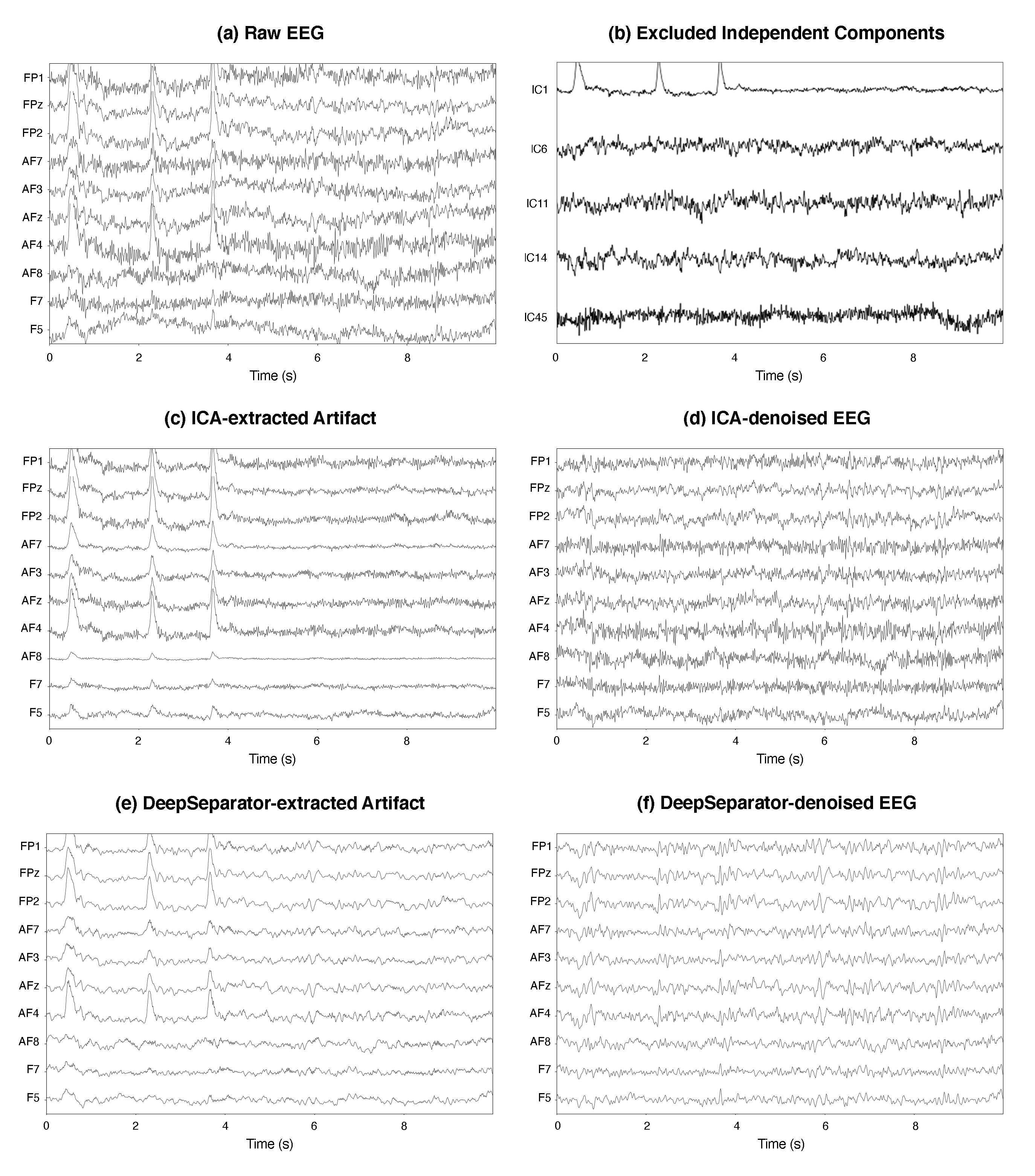}
\caption{\re{Performance on real multi-channel EEG signals: the time course of raw EEG signals (a), 5 excluded independent components decomposed by ICA (b), the artifacts extracted by ICA (c), the EEG signals denoised by ICA (d), the artifacts extracted by DeepSeparator (e), and the EEG signals denoised by DeepSeparator (f).
Both ICA and DeepSeparator can detect, extract and suppress the EOG artifact effectively. Simultaneously, DeepSeparator suppress the high-frequency artifacts which might be related to the EMG artifacts.}}
\label{fig:EEG_plot}
\end{figure*}

To visualize the denoising effects of ICA and DeepSeparator, we display the representative 10 channels in frontal lobe, where the EOG artifact is the mostly prominent (in Fig.~\ref{fig:EEG_plot}). It is obvious that the raw EEG contains EOG artifacts. Both DeepSeparator and ICA can effectively suppress the EOG peaks. Compared with ICA, there are two advantages of DeepSeparator: (i) DeepSeparator is fully automatic, without the necessity of manual selection of artifactual ICs; (ii) DeepSeparator implicitly learns to suppress the high-frequency artifacts which might be related to the EMG artifacts.

\subsubsection{ERP analysis}
The effects of EEG denoising might influence the downstream analysis, such as the ERP analysis and EEG source analysis. Here we investigate to what extent the DeepSeparaor-based EEG denoising influences the ERP analysis. We epoch the EEG time course of left auditory stimulation from 100ms prior to 400ms post stimulation, averaging over epochs to obtain the ERP.

\begin{figure*}[htb]
\centering
\includegraphics[width=1\textwidth]{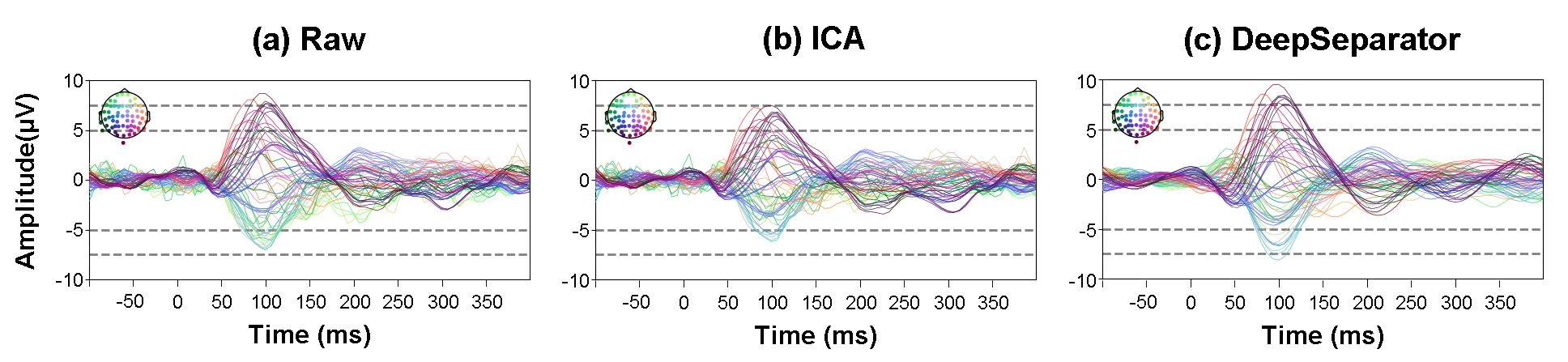}
\caption{Comparisons of ERP time courses: ERPs obtained from the raw EEG signals (a), the signals denoised by ICA (b) and DeepSeparator (c).}
\label{fig:ERP}
\end{figure*}

\begin{figure}[htb]
\centering
\includegraphics[width=0.6\textwidth]{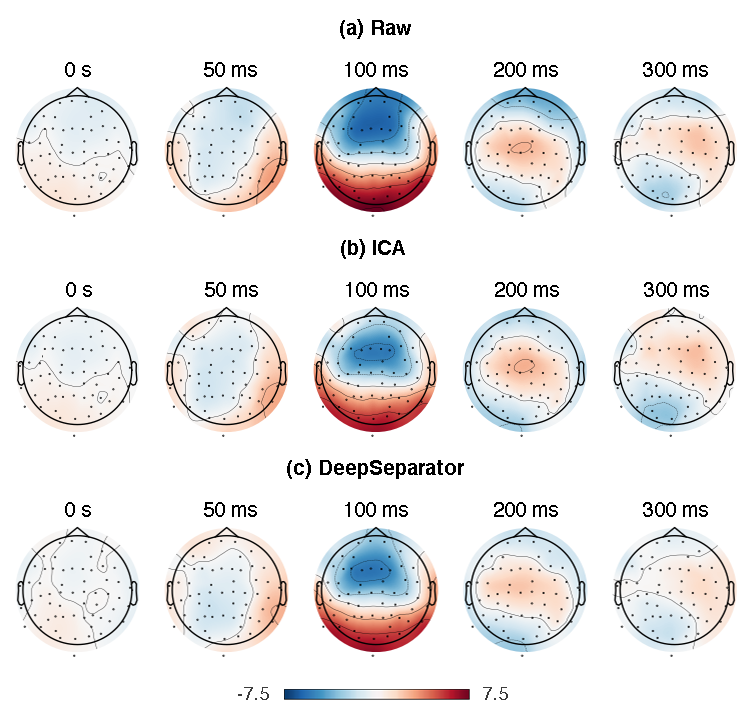}
\caption{Comparisons of ERP topography: ERPs obtained from the raw EEG signals (a), the signals denoised by ICA (b) and DeepSeparator (c). We present the topographic maps at the auditory stimulus onset (0s), as well as 50ms, 100ms, 200 post auditory stimulus onset. }
\label{fig:topo}
\end{figure}

The ERP time course from the raw EEG, ICA-denoised EEG and DeepSeparator-denoised EEG are shown in Fig.~\ref{fig:ERP}. The topography of ERP components at 0s, 50ms, 100ms, 200ms and 300ms are shown in Fig. \ref{fig:topo}. The results show that the ERP in the 100ms is weaker after ICA denoising, compared with the raw EEG and denoised EEG by DeepSeparator.  The right temporal areas are activated at 50ms, and the neural response to auditory stimulus is peaked at 100ms, which are well in line with previous auditory ERP study~\cite{hamalainen2007n1, korzyukov2012erp, hovel2015auditory}. 

\begin{figure*}[htb]
\centering
\includegraphics[width=0.9\textwidth]{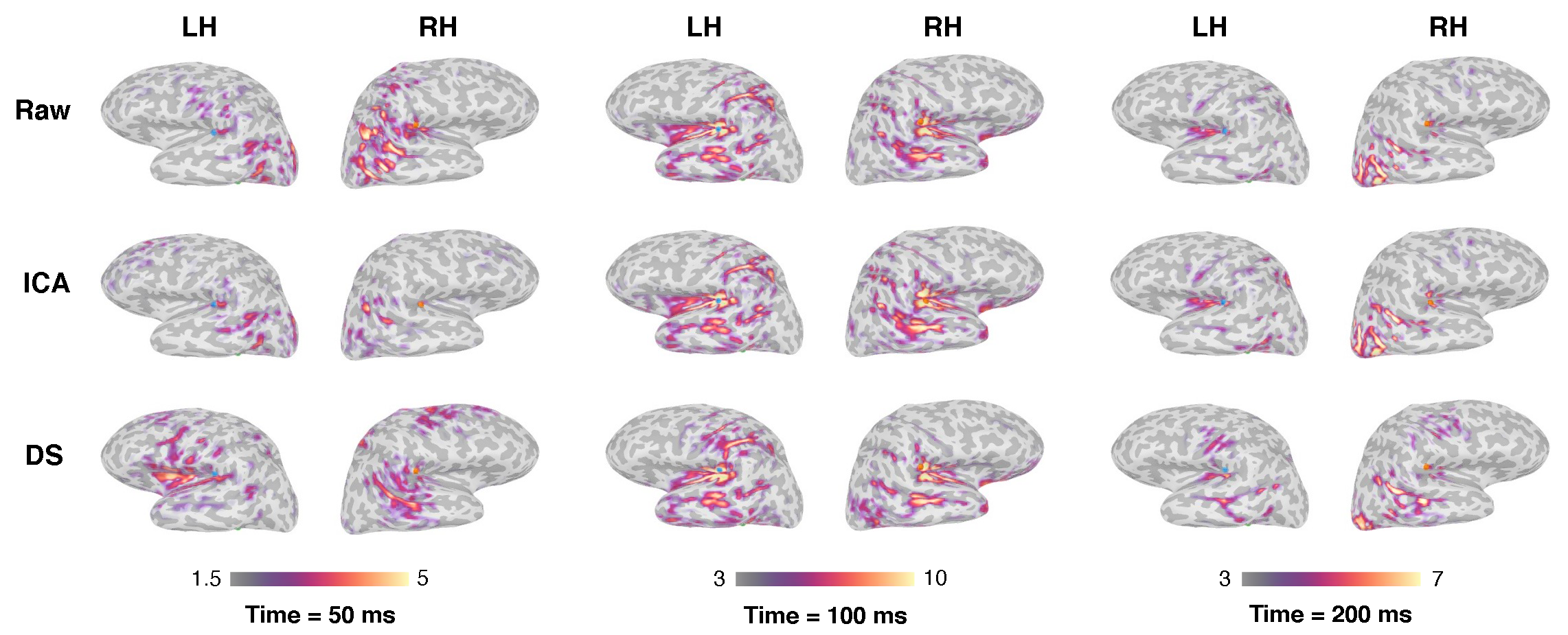}
\caption{Source localization results of the Raw signal, ICA denoised signal and DeepSeparator denoised signal.}
\label{fig:source}
\end{figure*}


\subsubsection{Source localization}

To further investigate the denoising effect on EEG source localization, we reconstruct the EEG sources with the raw EEG, ICA-denoised EEG, and DeepSeparator-denoised EEG. 
Specifically, we first build the head model for the forward problem, by using Boundary element method (BEM) on T1 image provided by the MNE-SAMPLE-DATASET~\cite{ds000248:1.2.1}. And then we apply dynamic statistical parametric maps (dSPM)~\cite{dale2000dynamic} to solve the inverse problem of EEG source localization, where the noise covariance is computed based on the signal baseline(e.g. 300ms prior the stimuli onset). We mainly examine the source activities at 50ms, 100ms and 200ms post stimuli onset, which are considered to be corresponding to the brainstem auditory evoked responses, midlatency responses and auditory long-latency sensory responses~\cite{pratt2010comparison}. 

The source maps at 50ms, 100ms and 200ms post stimuli from the raw EEG, the ICA-denoised EEG and DeepSeparator-denoised EEG are illustrated in Fig.~\ref{fig:source}. The main difference is that a weak response in the left insula is restored by DeepSeparator at 50ms, while the left insula activity is not visible in the sources from the raw EEG and ICA-denoised EEG. Previous study has indicated that insula is related to auditory processing ~\cite{bamiou2003insula,engelien1995functional,fifer1993insular,habib1995mutism}, suggesting the positive effects of DeepSeparator based EEG denoising on highlighting the neural sources.

\begin{figure}[htb]
\centering
\includegraphics[width=0.45\textwidth]{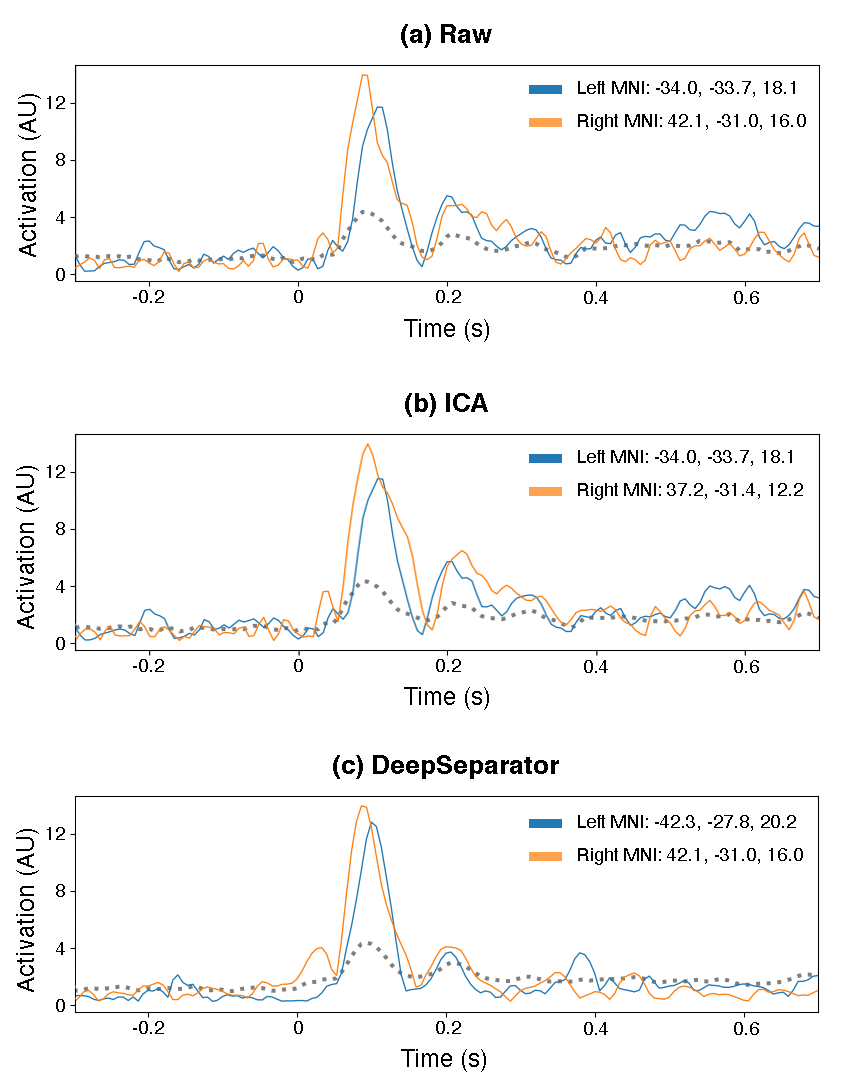}
\caption{Activity during stimulation of the voxel with highest activation. Compared with raw signal and signal processed by ICA, activation of signal processed by DeepSeparator is suppressed before stimulation and 400ms after stimulation. Neural activity is more significant from 50ms to 200ms.}
\label{fig:source_activation}
\end{figure}

The source dynamics at the peaking source are presented in Fig.~\ref{fig:source_activation}. The peaking sources of left hemisphere are all located in left supramarginal gyrus. The MNI coordinates of the source peak are [-34.0, -33.7, 18.1] for the raw EEG, [-34.0, -37.7, 18.1] for ICA-denoised EEG, [-42.3, -27.8, 20.2] for DeepSeparator-denoised EEG. As for right hemisphere, peaking sources of raw EEG and DeepSeparator-denoised EEG are located in right supramarginal gyrus, but the peaking source of ICA-denoised EEG is located in right primary auditory cortex. The MNI coordinates of the source peak are [42.1, -31.0, 16.0] for the raw EEG, [37.2, -31.4, 12.2] for ICA-denoised EEG, [42.1, -31.0, 16.0] for DeepSeparator-denoised EEG.
DeepSeparator can reduce the baseline ERP source activity (prior stimuli onset) which are considered as the noise, as well as reducing the leftover source activity (400ms after stimuli onset) which are expected to return to the baseline.



Overall, although DeepSeparator is trained with the semi-synthetic, single-channel EEG data, DeepSeparator is capable of transferring to the real EEG data and removes artifacts in multi-channel EEG surprisingly well, at least as well as ICA, if not better (Fig.~\ref{fig:EEG_plot}-\ref{fig:source_activation}).


\section{Discussion}
In this study, we proposed a deep learning model called DeepSeparator, which can extract the neural signal and the artifact from raw EEG data.

\subsubsection{From methodology perspective}
Removing artifacts mixed in the EEG signals is an important step for the following EEG analysis, such ERP analysis and source localization. However, previous methods, such as adaptive filtering and ICA, greatly rely on prior knowledge of the property of specific artifacts~\cite{correa2007artifact, wang2015removal, marino2018adaptive, nam2002independent,urrestarazu2004independent, de2006canonical}. As a deep learning model, DeepSeparator can achieve favorable performance in artifact removal tasks, especially for EOG artifact and EMG artifact, serving as a universal tool for removing various artifacts mixed in EEG. 

Theoretically, DeepSeparator can be considered as a nonlinear decomposition and reconstruction of the input, similar as the linear decomposition and reconstruction in ICA. 
Compared with ICA, the nonlinearity of DeepSeparator allows to learns more complex latent representations, and therefore higher capability of extract artifactual components. Moreover, DeepSeparator is well suited with single-channel EEG signal, while ICA can only work with multi-channel EEG signals.

Another merit of DeepSeparator is its interpretability, compared with other deep learning models. We analyzed the latent space of DeepSeparator in Fig.\ref{fig:visulization}, indicating that the encoder is responsible for capturing and amplifying the features in the raw signal, the decomposer for detecting and suppressing the artifacts in the embedding space, and the decoder for reconstructing the EEG signal and artifact based on the output from encoder and decomposer. The end-to-end architecture design (see Fig.\ref{fig:framework}) and the proposed training strategy (see Sec.~\ref{sec:training}) together allow to train and test DeepSeparator without external interventions; therefore the prior knowledge is not necessarily required for DeepSeparator.


\subsubsection{From application perspective}
Compared to previous methods which are designed for a specific type of artifacts or rely on prior knowledge for analysis~\cite{correa2007artifact, wang2015removal, marino2018adaptive, nam2002independent,urrestarazu2004independent, de2006canonical, zeng2015eemd, chen2018novel}, DeepSeparator can achieve satisfactory performance with single-channel or multi-channel EEG data for either EOG or EMG artifacts. This strong flexibility of DeepSeparator provides a great potential for applications on EEG denoising, such as for ERP analysis, source localization and EEG super resolution~\cite{gandhi2011eeg, alhaddad2014spectral, yu2012spatio, ivannikov2009erp, hild2009source}. Moreover, the performance of DeepSeparator is reliable at different SNR levels (Fig.~\ref{fig:snr_eog}\&\ref{fig:snr_emg}), which enables the use of DeepSeparator in noisy data. Once DeepSeparator has been trained, the running is fast and automatic. It can be further combined with the AI hardware for real-time EEG noise reduction, which would largely facilitate the EEG-based BCI applications.

\subsubsection{Limitation and future works} 
It is worth mentioning the limitations of our work.
First, DeepSeparator is not robust to the scale of the input. How to keep the denoising result and the input signal at the same scale is an important application issue.
Second, our model tends to discard high-frequency information (Fig~\ref{fig:EMG_freq}), which is a general issue of neural networks. As the Frequency principle of deep neural networks shown, neural networks tend to fit data from low frequency to high frequency~\cite{xu2019frequency}, resulting a bias towards low-dimensional features.
In our case, the EMG artifacts contaminate the high-frequency bands of EEG signals. It is more challenging to remove EMG artifacts compared to the low-frequency EOG artifacts.
\re{Besides, there are stereotypical artifacts (eg., ECG, EOG, EMG) and non-stereotypical artifacts that emerge due to uncontrolled motions of the subjects/patients or due to electrical discontinuities in EEG, DeepSeparator can remove stereotypical artifacts after training, but how to remove non-stereotypical artifacts is still a challenge for DeepSeparator.}
In addition, as a supervised deep learning model, the training process of DeepSeparator requires lots of expert-annotated data. \re{More verifications with expert-annotated datasets are important}, and unsupervised DL models are also our future direction~\cite{devlin2018bert, brown2020language, radford2019language}.
Finally, DeepSeparator does not use the complex spatial information across EEG channels; in contract, it processes each EEG channel separately. We believe that it is very important to design a model considering the spatio-temporal relationship between EEG channels, such as the graph networks~\cite{pentari2021graph}.

\section{Conclusion}
Artifact removal is an important topic for EEG analysis. In this study, we proposed a framework called DeepSeparator and verified its superiority over traditional approaches for EEG denoising. It can extract neural signals and artifacts from raw EEG of any number of channels at any length and may extend to other signal denoising problems.

\section*{Data availability statement}
The data and code that support the findings of this study are openly available at the following URL: \url{https://github.com/ncclabsustech/DeepSeparator}. \\
The semi-synthetic benchmark data, EEGdenoiseNET, is available at \url{https://github.com/ncclabsustech/EEGdenoiseNet}.

\section*{Acknowledgement}

We thank the anonymous reviewers for the insightful suggestions, and Mr. Haoming Zhang for his EEGdenoiseNET. This work was funded in part by the National Natural Science Foundation of China (62001205), Guangdong Natural Science Foundation Joint Fund (2019A1515111038), Shenzhen Science and Technology Innovation Committee (20200925155957004, KCXFZ2020122117340001, SGDX2020110309280100), Shenzhen Key Laboratory of Smart Healthcare Engineering (ZDSYS20200811144003009).

\section*{Conflict of interests}
All authors declare no competing interests.


\newpage
\newcommand{\newblock}{}

\bibliography{reference.bib}
\bibliographystyle{unsrt}

\end{document}